\documentclass[preprint]{elsarticle}

\usepackage{lineno,hyperref}
\modulolinenumbers[5]

\usepackage{microtype}
\usepackage{graphicx}
\usepackage{subfigure,xcolor}
\usepackage{booktabs} 
\usepackage{amssymb,amsmath,enumitem,caption,algorithmic,algorithm}
\hypersetup{hidelinks, colorlinks=true, linkcolor=blue,  anchorcolor=blue, citecolor=blue, filecolor=blue, menucolor=blue, urlcolor=blue}

\begin{document}

\begin{frontmatter}

\title{Bayesian Optimization with \\ Output-Weighted Optimal Sampling}

\author[mymainaddress]{Antoine Blanchard}
\cortext[mycorrespondingauthor]{Corresponding author}
\ead{ablancha@mit.edu}

\author[mymainaddress]{Themistoklis Sapsis\corref{mycorrespondingauthor}}
\ead{sapsis@mit.edu}

\address[mymainaddress]{Department of Mechanical Engineering, \\ Massachusetts Institute of Technology, Cambridge, MA 02139}

\begin{abstract}
In Bayesian optimization, accounting for the importance of the output relative to the input is a crucial yet challenging exercise, as it can considerably improve the final result but often involves inaccurate and cumbersome entropy estimations.  We approach the problem from the perspective of importance-sampling theory, and advocate the use of the likelihood ratio to guide the search algorithm towards regions of the input space where the objective function to be minimized assumes abnormally small values.  The likelihood ratio acts as a sampling weight and can be computed at each iteration without severely deteriorating the overall efficiency of the algorithm.  In particular, it can be approximated in a way that makes the approach tractable in high dimensions.  The ``likelihood-weighted'' acquisition functions introduced in this work are found to outperform their unweighted counterparts in a number of applications.   
\end{abstract}


\end{frontmatter}


\section{Introduction}
\label{sec:1}

The vast majority of optimization problems encountered in practical applications involve objective functions whose level of complexity prohibits use of classical optimization algorithms such as grid search, random search, and gradient-based methods (including Newton methods and conjugate gradient methods) \cite{jones1998efficient,hennig2012entropy}.  In fact, so little is known about the internal workings of those systems that practitioners have no choice but to treat them as ``black boxes'', for which a high evaluation cost adds to the issue of opacity, making optimization a daunting endeavor.

To solve such optimization problems while keeping the number of black-box evaluations at a minimum, one possibility is to use an iterative approach.  In this area, Bayesian optimization has received a great deal of attention because of its ability to a) incorporate prior belief one may have about the black-box objective function, and b) explore the input space carefully, compromising between exploration and exploitation before each function evaluation \cite{brochu2010tutorial,shahriari2015taking}.

A key component in Bayesian optimization lies in the choice of acquisition function, which is the ultimate decider as it commands where to next query the objective function.  Acquisition functions come in many shapes and forms, ranging from traditional improvement-based \cite{kushner1964new,mockus1978application,jones1998efficient} and optimistic policies \cite{srinivas2009gaussian,kaufmann2012bayesian} to more recent information-based strategies \cite{villemonteix2009informational,hennig2012entropy,hernandez2014predictive}.  The latter differ from the former in that they account for the importance of the output relative to the input (usually through the entropy of the posterior distribution over the unknown minimizer), leading to significant gains when the objective function is noisy and highly multi-modal.  

Despite superior empirical performance, information-based acquisition functions suffer from several shortcomings, including slow evaluation caused by heavy sampling requirements, laborious implementation, intractability in high dimensions, and limited choice of Gaussian process kernels \cite{hernandez2014predictive,hoffman2015output,wang2017max}.  The only exception of which we are aware is the fast information-theoretic Bayesian optimization (FITBO) algorithm \cite{ru2017fast}, in which efficiency and flexibility come at the cost of a decline in performance, the latter being just ``on par'' with traditional improvement-based and optimistic policies.

While it is clear that incorporating information about the output space in an acquisition function has significant merit, doing so while eliminating the above limitations calls for a different approach.  Inspired by the theory of importance sampling \cite{owen2013monte}, we propose to equip acquisition functions with the \emph{likelihood ratio}, a quantity that accounts for the importance of the output relative to the input and which appears frequently in uncertainty quantification and experimental design problems \cite{chaloner1995bayesian,huan2013simulation}.  The significance of the proposed approach is manifold:

\begin{enumerate}

\item The likelihood ratio acts as a probabilistic sampling weight and guides the algorithm towards regions of the input space where the objective function assumes abnormal values, which is beneficial in cases where the minima of the objective function correspond to rare and extreme output values.

\item In addition to output information, the likelihood ratio encapsulates any prior knowledge one may have about the distribution of the input, which has significant implications in problems related to uncertainty quantification and experimental design where the input space often comes equipped with a non-uniform probability distribution.

\item The likelihood ratio can be satisfactorily approximated with a Gaussian mixture model, making it possible to compute any ``likelihood-weighted'' acquisition function (and its gradients) analytically for a range of Gaussian process kernels and therefore allowing for the possibility of the input space being high-dimensional.
\end{enumerate}

Before going further, a word of caution is in order regarding the alleged advantages of the proposed approach.  Anyone engaged in solving black-box optimization problem must be mindful of the no-free-lunch theorem which states that ``if an algorithm performs well on a certain class of problems then it necessarily pays for that with degraded performance on the set of all remaining problems'' \cite{wolpert1997no}.  Our approach is no exception to the rule.  It is expected to provide an advantage in situations where a) Bayesian optimization is a suitable candidate, and b) the global minimum of the objective function is ``extreme'', that is, it is separated from the remainder of the optimization landscape by several standard deviations.  In all other situations, the proposed approach may provide no advantage at all, and even perform worse than other algorithms such as genetic programming or random-restart hill climbing.

\section{Formulation of the Problem}
\label{sec:2}

\subsection{A Brief Review of Bayesian Optimization}
\label{sec:21}

We consider the problem of finding a global minimizer of a function $f : \mathbb{R}^d \longrightarrow \mathbb{R}$ over a compact set $\mathcal{X} \subset \mathbb{R}^d$.  This is usually written as
\begin{equation}
\min_{\mathbf{x} \in \mathcal{X}} f(\mathbf{x}).
\label{eq:1}
\end{equation}
We assume that the objective function $f$ is unknown and therefore treat it as a black box.  In words, this means that $f$ has no simple closed form, and neither do its gradients.  This allows for the possibility of $f$ being nonlinear, non-convex and multi-peak, although we do require that $f$ be Lipschitz continuous to avoid pathological cases \cite{brochu2010tutorial}.  The objective function can, however, be evaluated at any arbitrary query point $\mathbf{x} \in \mathcal{X}$, each evaluation producing a potentially noise-corrupted output $y \in \mathbb{R}$.  In this work, we model uncertainty in observations with additive Gaussian noise:
\begin{equation}
y = f(\mathbf{x}) + \varepsilon, \quad \varepsilon \sim \mathcal{N}(0, \sigma_\varepsilon^2).
\end{equation}
In practice, the function $f$ can be a machine-learning algorithm (with $\mathbf{x}$ the hyper-parameters), a large-scale computer simulation of a physical system (with $\mathbf{x}$ the physical parameters), or a field experiment (with $\mathbf{x}$ the experimental parameters).  Evaluating $f$ can thus be very costly, so to solve the minimization problem \eqref{eq:1} each query point must be selected very meticulously.

Bayesian optimization is a sequential approach which, starting from an initial dataset of input--output pairs, iteratively probes the input space and, with each point visited, attempts to construct a surrogate model for the objective function.  At each iteration, the ``best next point'' to visit is determined by minimizing an acquisition function $a : \mathbb{R}^d \longrightarrow \mathbb{R}$ which serves as a guide for the algorithm as it explores the input space.  After a specified number of iterations, the algorithm uses the surrogate model it has constructed to make a final recommendation for what it believes is the true minimizer of the objective function (Algorithm \ref{alg:1}).

\begin{algorithm}[tb]
   \caption{Bayesian optimization}
   \label{alg:1}
\begin{algorithmic}
   \STATE {\bfseries Input:} Number of initial points $n_\textit{init}$, number of iterations $n_\textit{iter}$ 
   \STATE {\bfseries Initialize:} Surrogate model $\bar{f}$ on initial dataset $\mathcal{D}_0 = \{\mathbf{x}_i, y_i\}_{i=1}^{n_\textit{init}}$
   \FOR{$n=0$ {\bfseries to} $n_\textit{iter}$}
      \STATE Select best next point $\mathbf{x}_{n+1}$ by minimizing acquisition function:
      	\begin{equation*}
	\mathbf{x}_{n+1} = \operatorname*{arg\,min}_{\mathbf{x} \in \mathcal{X}} a(\mathbf{x}; \bar{f}, \mathcal{D}_n)
	\end{equation*}
	\STATE Evaluate objective function $f$ at $\mathbf{x}_{n+1}$ and record $y_{n+1}$
	\STATE Augment dataset: $\mathcal{D}_{n+1} = \mathcal{D}_n \cup \{\mathbf{x}_{n+1} , y_{n+1} \}$
   \STATE Update surrogate model
   \ENDFOR
   \STATE \textbf{Return:} Final recommendation from surrogate model
         \begin{equation*}
	\mathbf{x}^* = \operatorname*{arg\,min}_{\mathbf{x} \in \mathcal{X}} \bar{f}(\mathbf{x})
	\end{equation*}
\end{algorithmic}
\end{algorithm}

The two key issues in Bayesian optimization are the choice of surrogate model $\bar{f}$ and the choice of acquisition function $a$.  The former is important because it represents our belief about what the objective function looks like given the data collected by the algorithm; the latter is important because it guides the algorithm in its exploration of the input space.  For Bayesian optimization to provide any sort of advantage over a brute-force approach, the costs of constructing the surrogate model and optimizing the acquisition function must be small compared to that of evaluating the black-box function $f$.

\subsection{Model Selection}
\label{sec:22}

Many different surrogate models have been used in Bayesian optimization, with various levels of success \cite{shahriari2015taking}.  In this work, we use a non-parametric Bayesian approach based on Gaussian process (GP) regression \cite{rasmussen2006gaussian}.  This choice is appropriate for Bayesian optimization because Gaussian processes a) are agnostic to the details of the black box, and b) provide a way to quantify uncertainty associated with noisy observations \cite{raissi2017inferring,raissi2017machine,pang2017discovering}.

A Gaussian process $\bar{f}(\mathbf{x})$ is completely specified by its mean function $m(\mathbf{x})$ and covariance function $k(\mathbf{x},\mathbf{x}')$.  For a dataset $\mathcal{D}$ of input--output pairs (written in matrix form as $\{\mathbf{X}, \mathbf{y}\}$) and a Gaussian process with constant mean $m_0$, the random process $\bar{f}(\mathbf{x})$ conditioned on $\mathcal{D}$ follows a normal distribution with posterior mean and variance
\begin{subequations}
\begin{gather}
\mu(\mathbf{x}) = m_0 + k(\mathbf{x}, \mathbf{X}) \mathbf{K}^{-1} (\mathbf{y} -m_0), \label{eq:3a}\\
\sigma^2(\mathbf{x}) = k(\mathbf{x},\mathbf{x}) - k(\mathbf{x}, \mathbf{X}) \mathbf{K}^{-1} k(\mathbf{X},\mathbf{x}),\label{eq:3b}
\end{gather}
\end{subequations}
respectively, where $\mathbf{K} = k(\mathbf{X},\mathbf{X}) + \sigma_\varepsilon^2 \mathbf{I}$.  Equation \eqref{eq:3a} can be used to predict the value of the surrogate model at any point $\mathbf{x}$, and \eqref{eq:3b} to quantify uncertainty in prediction at that point \cite{rasmussen2006gaussian}.  

In GP regression, the choice of covariance function is crucial, and in what follows we use the radial-basis-function (RBF) kernel with automatic relevance determination (ARD):
\begin{equation}
k(\mathbf{x},\mathbf{x}') = \sigma_f^2 \exp \!\left[ -(\mathbf{x} - \mathbf{x}')^\mathsf{T} \mathbf{\Theta}^{-1}(\mathbf{x} - \mathbf{x}') /2\right],
\end{equation}
where $\mathbf{\Theta}$ is a diagonal matrix containing the lengthscales for each dimension.  Other choices of kernels are possible (e.g., linear kernel or Mat\'ern kernel), but we will see that the RBF function has several desirable properties that will come in handy when designing our algorithm.

Exact inference in Gaussian process regression requires inverting the matrix $\mathbf{K}$, typically at each iteration.  This is usually done by Cholesky decomposition whose cost scales like $O(n^3)$, with $n$ the number of observations \cite{rasmussen2006gaussian}.  (A cost of $O(n^2)$ can be achieved if the parameters of the covariance function are fixed.)  Although an $O(n^3)$ scaling may seem daunting, it is important to note that in Bayesian optimization the number of observations (i.e., function evaluations) rarely exceeds a few dozens (or perhaps a few hundreds), as an unreasonably large number of observations would defeat the whole purpose of the algorithm.

\subsection{Acquisition Functions for Bayesian Optimization}
\label{sec:23}

The acquisition function is at the core of the Bayesian optimization algorithm, as it solely determines the points at which to query the objective function.  The role of the acquisition function is to find a compromise between exploration (i.e., visiting regions where uncertainty is high) and exploitation (i.e., visiting regions where the surrogate model predicts small values).  In this work, we consider the following three classical acquisition functions.

\textbf{Probability of Improvement (PI).} Given the current best observation $y^*$, PI attempts to maximize 
\begin{equation}
a_\textit{PI}(\mathbf{x}) =\Phi(\lambda(\mathbf{x})),
\label{eq:pi}
\end{equation}
where $\lambda(\mathbf{x}) = [y^* - \mu(\mathbf{x}) - \xi]/\sigma(\mathbf{x})$, $\Phi$ is the cumulative density function of the standard normal distribution, and $\xi \geq 0$ is a user-defined parameter that controls the trade-off between exploration and exploitation \citep{jones1998efficient}.  

\textbf{Expected Improvement (EI).} Arguably the most popular acquisition function for Bayesian optimization, EI improves on PI in that it also accounts for \textit{how much} improvement a point can potentially yield \citep{jones1998efficient}:
\begin{equation}
a_\textit{EI}(\mathbf{x}) =\sigma(\mathbf{x}) \left[\lambda(\mathbf{x}) \Phi(\lambda(\mathbf{x})) + \phi(\lambda(\mathbf{x})) \right],
\label{eq:ei}
\end{equation}
where $\phi$ is the probability density function (pdf) of the standard normal distribution.

\textbf{Lower Confidence Bound (LCB).}  In LCB, the best next point is selected based on the bandit strategy of \citet{srinivas2009gaussian}:
\begin{equation}
a_\textit{LCB}(\mathbf{x}) = \mu(\mathbf{x}) - \kappa \sigma(\mathbf{x}),
\label{eq:7}
\end{equation}
where $\kappa\geq0$ is a user-specified parameter that balances exploration (large $\kappa$) and exploitation (small $\kappa$).  

The popularity of PI, EI and LCB can be largely explained by the facts that a)  implementation is straightforward, b) evaluation is inexpensive, and c) gradients are readily accessible, opening the door for gradient-based optimizers.  The combination of these features makes the search for the best next point considerably more efficient than otherwise.  

It is important to note that the rationale behind LCB is to ``repurpose'' a purely explorative acquisition function which reduces uncertainty globally, $\sigma(\mathbf{x})$, into one suitable for Bayesian optimization which is more aggressive towards minima.  This is done by appending $\mu(\mathbf{x})$ and introducing the trade-off parameter $\kappa$, as done in \eqref{eq:7}.  Following the same logic, we repurpose the Integrated Variance Reduction (IVR) of \citet{sacks1989design},
\begin{equation}
a_\textit{IVR}(\mathbf{x}) = \frac{1}{\sigma^2(\mathbf{x})} \int \mathrm{cov}^2(\mathbf{x},\mathbf{x}')\, \mathrm{d}\mathbf{x}',
\label{eq:8}
\end{equation}
into
\begin{equation}
a_\textit{IVR-BO}(\mathbf{x}) = \mu(\mathbf{x}) -  \kappa\, a_\textit{IVR}(\mathbf{x}),
\label{eq:9}
\end{equation}
where the suffix ``BO'' in ``IVR-BO'' stands for ``Bayesian Optimization''.  In the above, $\mathrm{cov}(\mathbf{x},\mathbf{x}')$ denotes the posterior covariance between $\mathbf{x}$ and $\mathbf{x}'$.  The formula for IVR involves an integral over the input space, but for the RBF kernel this integral can be computed analytically, and the same is true of its gradients (\ref{app:1}).  Therefore, IVR-BO retains the three important features discussed earlier for PI, EI and LCB.  

\section{Bayesian Optimization with Output-Weighted Optimal Sampling}
\label{sec:3}

In this section, we introduce a number of novel acquisition functions for Bayesian optimization, each having the following features: a) they leverage previously collected information by assigning more weight to regions of the search space where the objective function assumes extreme values;  b) they incorporate a prior $p_\mathbf{x}$ over the search space, allowing the algorithm to focus on regions of potential relevance; and c) their computational complexity does not negatively affect the overall efficiency of the sequential algorithm.  

\subsection{Likelihood-Weighted Acquisition Functions}
\label{sec:31}

To construct these novel acquisition functions, we proceed in two steps.  First, we introduce a purely explorative acquisition function that focuses on abnormal output values without any distinction being made between maxima and minima.  Then, we repurpose this acquisition function as was done for LCB and IVR-BO.  The repurposed criterion is suitable for Bayesian optimization as it specifically targets abnormally small output values (i.e., extreme minima).

When the objective function assumes rare and extreme (i.e., abnormally large or small) output values, then the conditional pdf of the output $p_{f|\mathbf{x}}$ is heavy-tailed.  (A pdf is heavy-tailed when at least one of its tails is not exponentially bounded.)  Heavy-tailed distributions are the manifestation of high-impact events occurring with low probability, and as a result they commonly arise in the study of risk \citep{embrechts2013modelling} and extreme events \citep{albeverio2006extreme}.  

One possible strategy, therefore, is for the sequential algorithm to use the pdf of the GP mean $p_{\mu}$ as a proxy for $p_{f}$ and select the best next point so that uncertainty in $p_{\mu}$ is most reduced.  The latter can be quantified by
\begin{equation}
a_L(\mathbf{x}) = \int \left| \log p_{\mu_+}(y) - \log p_{\mu_-}(y)  \right| \mathrm{d}y,
\label{eq:10}
\end{equation}
where $\mu_{\pm}(\mathbf{x}'; \mathbf{x})$ denotes the upper and lower confidence bounds at $\mathbf{x}'$ had the data point $\{\mathbf{x}, \mu(\mathbf{x})\}$ been collected, that is, $\mu_{\pm}(\mathbf{x}'; \mathbf{x}) =  \mu(\mathbf{x}') \pm \sigma^2(\mathbf{x}'; \mathbf{x})$.  The use of logarithms in \eqref{eq:10} places extra emphasis on the pdf tails where extreme minima and maxima ``live'' \citep{mohamad2018sequential}.

The above metric enjoys attractive convergence properties \citep{mohamad2018sequential} but is cumbersome to compute (not to mention optimize) and intractable in high dimensions.  So instead we show that $a_L(\mathbf{x})$ is bounded above (up to a multiplicative constant) by 
\begin{equation}
a_B(\mathbf{x}) = \int \sigma^2(\mathbf{x}'; \mathbf{x})  \frac{p_\mathbf{x}(\mathbf{x}')}{p_\mu(\mu(\mathbf{x}'))} \, \mathrm{d}\mathbf{x}'.
\label{eq:11}
\end{equation}
Equation \eqref{eq:11} is a massive improvement over \eqref{eq:10} from the standpoint of reducing complexity.  More importantly, it reveals an unexpected connection between the metric $a_L$ (whose primary focus is the reduction of uncertainty in pdf tails) and IVR acquisition function in \eqref{eq:8}.  Indeed, it only takes a few lines to show that $a_B(\mathbf{x})$ is strictly equivalent to 
\begin{equation}
a_\textit{IVR-LW}(\mathbf{x}) = \frac{1}{\sigma^2(\mathbf{x})} \int \mathrm{cov}^2(\mathbf{x}, \mathbf{x}') \frac{p_\mathbf{x}(\mathbf{x}')}{p_\mu(\mu(\mathbf{x}'))} \, \mathrm{d}\mathbf{x}',
\label{eq:12}
\end{equation}
which is clearly a cousin of \eqref{eq:8}, with the ratio $p_\mathbf{x}(\mathbf{x})/p_\mu(\mu(\mathbf{x}))$ playing the role of a \emph{sampling weight} (\ref{app:2}).  The suffix ``LW'' in ``IVR-LW'' stands for ``likelihood-weighted''.

With \eqref{eq:12} in hand, two remarks are in order.  First, we can establish the convergence of \eqref{eq:12} by recognizing that there exists a constant $M>0$ such that for all $\mathbf{x} \in \mathbb{R}^d$, 
\begin{equation}
0 \leq \frac{p_\mathbf{x}(\mathbf{x})}{p_\mu(\mu(\mathbf{x}))} \leq M.
\end{equation}
Therefore, we have that $0 \leq a_\textit{IVR-LW}(\mathbf{x}) \leq M a_\textit{IVR}(\mathbf{x})$; and since $a_\textit{IVR}$ goes to zero in the limit of many observations, so does $a_\textit{IVR-LW}$.  Second, it is important to emphasize that \eqref{eq:12} is a purely explorative acquisition function which does not discriminate between extreme minima (i.e., heavy left tail) and extreme maxima (i.e., heavy right tail).

To make IVR-LW suitable for Bayesian optimization, we repurpose \eqref{eq:12} as 
\begin{equation}
a_\textit{IVR-LWBO}(\mathbf{x}) = \mu(\mathbf{x}) - \kappa \,a_\textit{IVR-LW}(\mathbf{x}),
\label{eq:13}
\end{equation}
which specifically targets extreme minima, consistent with \eqref{eq:1}.  By the same logic, we also introduce
\begin{equation}
a_\textit{LCB-LW}(\mathbf{x}) = \mu(\mathbf{x}) - \kappa  \sigma(\mathbf{x}) \frac{p_\mathbf{x}(\mathbf{x})}{p_\mu(\mu(\mathbf{x}))} \label{eq:15}
\end{equation}
as the likelihood-weighted counterpart to LCB.  Here, it is natural to incorporate the ratio $p_\mathbf{x}(\mathbf{x})/p_\mu(\mu(\mathbf{x}))$ in a product with the explorative term $\sigma(\mathbf{x})$.  This is in the same spirit as \eqref{eq:11} where that ratio naturally appears as a sampling weight for the posterior variance.

\subsection{The Role of the Likelihood Ratio}
\label{sec:32}

In the importance-sampling literature, the ratio
\begin{equation}
w(\mathbf{x}) = \frac{p_\mathbf{x}(\mathbf{x})}{p_\mu(\mu(\mathbf{x}))}
\label{eq:14}
\end{equation}
is referred to as the \emph{likelihood ratio} \citep{owen2013monte}.  (In that context, the distributions $p_\mathbf{x}$ and $p_\mu$ are referred to as the ``nominal distribution'' and ``importance distribution'', respectively.)  The likelihood ratio is important in cases where some points are more important than others in determining the value of the output.  For points with similar probability of being observed ``in the wild'' (i.e., same $p_\mathbf{x}$), the likelihood ratio assigns more weight to those that have a large impact on the magnitude of output (i.e., small $p_\mu$).  For points with similar impact on the output (i.e., same $p_\mu$), it promotes those with higher probability of occurrence (i.e., large $p_\mathbf{x}$).  In other words, the likelihood ratio favors points for which the magnitude of the output is \textit{unusually large} over points associated with frequent, average output values.


In Bayesian optimization, the likelihood ratio can be beneficial in at least two ways.  First, it should improve performance in situations in which the global minima of the objective function ``live'' in the (heavy) left tail of the output pdf $p_f$.  Second, the likelihood ratio makes it possible to distill any prior knowledge one may have about the distribution of the input.  For ``vanilla'' Bayesian optimization, it is natural to use a uniform prior for $p_\mathbf{x}$ because in general any point $\mathbf{x}$ is as good as any other.  But if the optimization problem arises in the context of uncertainty quantification of a physical system (where generally the input space comes equipped with a non-uniform probability distribution), or if one has prior beliefs about where the global minimizer may be located, then use of a non-trivial $p_\mathbf{x}$ has the potential to considerably improve performance of the search algorithm.

As far as we know, use of a prior on the input space is virtually unheard of in Bayesian optimization, the only exceptions being the works of \citet{bergstra2011algorithms} and \citet{oliveira2019bayesian}.  There, a prior is placed on the input space in order to account for \emph{localization noise}, i.e., the error in estimating the query $\mathbf{x}$ location with good accuracy.  This is quite different from our approach in which the query locations are assumed to be known with exactitude and the input prior is used to highlight certain regions of the search space before the search is initiated.  We also note that in GP regression, prior beliefs about the objective function are generally encoded in the covariance function, and its selection is a delicate matter even for the experienced practitioner.  Using a prior on the input space may be viewed as a simple way of encoding structure without having to resort to convoluted GP kernels.   

\subsection{Approximation of the Likelihood Ratio}
\label{sec:33}

We must ensure that introduction of the likelihood ratio does not compromise our ability to compute the acquisition functions and their gradients efficiently.  We first note that to evaluate the likelihood ratio, we must estimate the conditional pdf of the posterior mean $p_\mu$, typically at each iteration.  This is done by computing $\mu(\mathbf{x})$ for a large number of input points and applying kernel density estimation (KDE) to the resulting samples.  Fortunately, KDE is to be performed in the one-dimensional output space, allowing use of fast FFT-based algorithms which scale linearly with the number of samples \cite{fan1994fast}.  

Yet, the issue remains that in IVR-LW(BO), the likelihood ratio appears in an integral over the input space.  To avoid resorting to Monte Carlo integration, we approximate $w(\mathbf{x})$ with a Gaussian mixture model (GMM):
\begin{equation}
w(\mathbf{x})  \approx \sum_{i=1}^{n_\textit{GMM}} \alpha_i \, \mathcal{N}(\mathbf{x}; \boldsymbol{\omega}_i, \mathbf{\Sigma}_i).
\label{eq:16}
\end{equation}
The GMM approximation has two advantages.  First, when combined with the RBF kernel, the problematic integrals and their gradients become analytic (\ref{app:3}).  This is important because it makes the approach tractable in high dimensions, a situation with which sampling-based acquisition functions are known to struggle.  Second, the GMM approximation imposes no restriction on the nature of $p_\mathbf{x}$ and $p_\mu$.  This is an improvement over the approach of \citet{sapsis2020output} whose approximation of $1/p_\mu$ by a second-order polynomial led to the requirement that $p_\mathbf{x}$ be Gaussian or uniform.  

The number of Gaussian mixtures to be used in \eqref{eq:16} is at the discretion of the user.  In this work, $n_\textit{GMM}$ is kept constant throughout the search, although the option exists to modify it ``on the fly'', either according to a predefined schedule or by selecting the value of $n_\textit{GMM}$ that minimizes the Akaike Information Criterion (AIC) or the Bayesian Information Criterion (BIC) at each iteration \cite{vanderplas2016python}.  We note that the optimal value of $n_\textit{GMM}$ might be unreasonably large for very high-dimensional systems.  Should that happen, we recommend prescribing a threshold value for $n_\textit{GMM}$ not to exceed.  The resulting mixture model will be suboptimal, but this is a small price to pay to alleviate the curse of dimensionality given that the extreme regions need not be localized with pinpoint accuracy.  (A coarse representation of these regions suffices.)

\begin{figure}[!htb]
\centering
\subfigure[2-D Ackley function]{\includegraphics[width=3in]{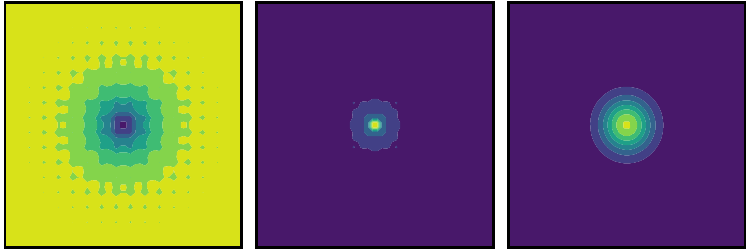}}

\subfigure[2-D Michalewicz function]{\includegraphics[width=3in]{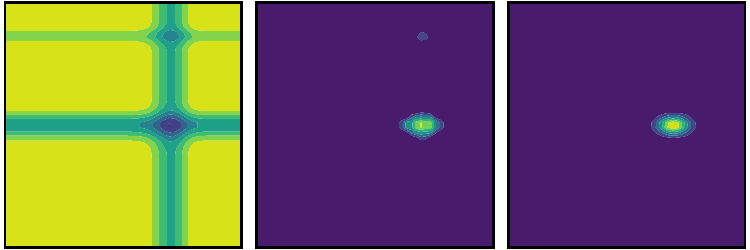}}

\caption{Contour plots of the objective function (left panel), the likelihood ratio with uniform $p_\mathbf{x}$ (center panel), and the GMM approximation (right panel).  We used $n_\textit{GMM}=2$ for the 2-D Ackley function and $n_\textit{GMM}=4$ for the 2-D Michalewicz function.}
\label{fig:1}
\end{figure}

We illustrate the benefits of using the likelihood ratio in the 2-D Ackley and Michalewicz functions, two test functions notoriously challenging for optimization algorithms.  (Analytical expressions are given in \ref{app:4}.)  For these functions, figure \ref{fig:1} makes it visually clear that the likelihood ratio gives more emphasis to the region where the global minimum is located (i.e., the center region for the Ackley function, and the slightly off-center region for the Michalewicz function).  Figure \ref{fig:1} also shows that $w(\mathbf{x})$ can be approximated satisfactorily with a small number of Gaussian mixtures, a key prerequisite for preserving algorithm efficiency.

We summarize our algorithm for computation of likelihood-weighted acquisition functions in Algorithm \ref{alg:2}.  Up to now we have adhered to conventional notation at the expense of making it clear whether the acquisition functions should be minimized or maximized.  The summary presented in Table \ref{sample-table} should dissipate any ambiguity.  

\begin{algorithm}[!htb]
   \caption{Likelihood-weighted acquisition function}
   \label{alg:2}
\begin{algorithmic}
   \STATE {\bfseries Input:} Posterior mean $\mu(\mathbf{x})$, input pdf $p_\mathbf{x}$, functional form for LW acquisition function $a(\mathbf{x}; w(\mathbf{x}))$, number of Gaussian mixtures $n_\textit{GMM}$

\algorithmicdo\ 
\begin{ALC@g}
     \STATE Sample posterior mean $\mu(\mathbf{x})$
     \STATE Estimate $p_\mu$ by KDE
     \STATE Compute  $w(\mathbf{x}) = p_\mathbf{x}(\mathbf{x})/p_\mu(\mathbf{x})$
     \STATE $w_\textit{GMM}(\mathbf{x})$ $\leftarrow$ Fit GMM to $w(\mathbf{x})$

\end{ALC@g}
\algorithmicend~\algorithmicdo\
   \STATE \textbf{Return:} $a(\mathbf{x}; w_\textit{GMM}(\mathbf{x}))$ and   gradients in analytic form \end{algorithmic}
\end{algorithm}

\begin{table}[!ht]
\caption{Summary of the acquisition functions considered in this work.}
\label{sample-table}
\vskip 0.15in
\begin{center}
\begin{small}
\begin{sc}
\begin{tabular}{lll}
\toprule
Acquisition Function & Equation & Rule  \\
\midrule
PI 			& (\ref{eq:pi}) 	& maximize \\
EI    			& (\ref{eq:ei}) 	& maximize  \\
LCB(-LW)    	&  (\ref{eq:7}), (\ref{eq:15})	& minimize  \\
IVR(-LW)    	& (\ref{eq:8}), (\ref{eq:12})	& maximize  \\
IVR-BO(LW)     & (\ref{eq:9}), (\ref{eq:13})	& minimize  \\
\bottomrule
\end{tabular}
\end{sc}
\end{small}
\end{center}
\vskip -0.1in
\end{table}

\section{Results}
\label{sec:4}

\subsection{Experimental protocol}
\label{sec:41}

To demonstrate the benefits of the likelihood ratio, we conduct a series of numerical experiments with the acquisition functions introduced in Section \ref{sec:3}.  Specifically, we compare EI, PI, LCB(-LW), IVR(-BO), and IVR-LW(BO).  We include IVR and IVR-LW in this list because they provide insight about the behavior of IVR-BO and IVR-LWBO, respectively, in the limit of large $\kappa$.  

For each example considered, we indicate the number of Gaussian mixtures used in the GMM approximation.  We use $\xi=0.01$ for EI and PI, and $\kappa=1$ for LCB, IVR and their respective variants.  (Although not considered in this work, use of a schedule for $\kappa$ has the potential to favorably affect the algorithm \cite{srinivas2009gaussian}.)  We do not \textit{set} the noise variance $\sigma_\varepsilon^2$, but rather let the algorithm learn it from data.

When the location of the global minimum $\mathbf{x}_\textit{true}$ and corresponding function value $y_\textit{true}$ are known (as in Section \ref{sec:42}), we report the simple regret
\begin{equation}
r(n) = \min_{k\in [0, n]} f(\mathbf{x}^*_k) - y_\textit{true},
\label{eq:17}
\end{equation}
where $\mathbf{x}^*_k$ denotes the optimizer recommendation at iteration $k$, and $n$ is the index of the current iteration \cite{wang2017max}.  Because several of the functions considered thereafter are highly multimodal and oscillatory, we also report the distance
\begin{equation}
\ell(n) = \min_{k\in [0, n]} \Vert \mathbf{x}_\textit{true} - \mathbf{x}^*_k \Vert^2,
\end{equation}
as in \citet{ru2017fast}.  (When the function has multiple global minimizers, we compute $\ell$ for each minimizer and report the smallest value.)  When the global minimizers of the objective function are not known a priori (as in Section \ref{sec:43}), we use \eqref{eq:17} with $y_\textit{true}=0$, which we complement with the observation regret
\begin{equation}
r_o(n) = \min_{y_i \in \mathcal{D}_n} y_i.
\end{equation}
For each example considered, we run 100 Bayesian experiments, each differing in the choice of initial points, and report the median for the metrics introduced above.  (The shaded error bands indicate a quarter of the median absolute deviation.)  Experiments were performed on a computer cluster equipped with 40 Intel Xeon E5-2630v4 processors clocking at 2.2 GHz.  Our code is available on GitHub.\footnote{\url{https://github.com/ablancha/gpsearch}}

\subsection{Benchmark of Test Functions}
\label{sec:42}

We begin with the benchmark of six test functions commonly used to evaluate performance of optimization algorithms (\ref{app:4}).  The Branin and 6-D Hartmann functions are moderately difficult, while the other four are particularly challenging:  the Ackley function features numerous local minima found in a nearly flat region surrounding the global minimum; the Bukin function is non-differentiable with a steep ridge filled with local minima; and the Michalewicz functions have steep valleys and ridges with a global minimum accounting for a tiny fraction of the search space.  

For each test function, the input space is rescaled to the unit hypercube so as to facilitate training of the GP hyper-parameters.  The noise variance is specified as $\sigma_\varepsilon^2=10^{-3}$ and appropriately rescaled to account for the variance of the objective function.  We use $n_\textit{init}=3$ for the 2-D functions and $n_\textit{init}=10$ otherwise.  In each case the initial points are selected by Latin hypercube sampling (LHS).

For uniform $p_\mathbf{x}$ and $n_\textit{GMM} = 2$, figure \ref{fig:2} shows that the LW acquisition functions systematically and substantially outperform their unweighted counterparts in identifying the location and objective value of the global minimum.  The only exception is with the Branin function, for which the likelihood ratio appears to have no positive effect.  For a possible explanation, consider that the output pdf of the Branin function has a very light left tail; by contrast, the other objective functions have heavy left tails (figure \ref{fig:S1}).  This observation is consistent with the derivation in Section \ref{sec:31} in which it was shown that the LW acquisition functions primarily target abnormally small output values, strongly suggesting that a heavy left tail is a prerequisite for LW acquisition functions to be competitive.

Figure \ref{fig:2} confirms that use of the likelihood ratio allows the algorithm to explore the search space more efficiently by focusing on regions where the magnitude of the objective function is thought to be unusually large.  For a clear manifestation of this, consider the left panel in figure \ref{fig:2c}, where the three LW acquisition functions dramatically outclass the competition, a consequence of the fact that the likelihood ratio helps the algorithm target the ridge of the Bukin function much more rapidly and thoroughly than otherwise.

Figures \ref{fig:2e} and \ref{fig:2f} demonstrate superiority of our approach in high dimensions.  Overall, figure \ref{fig:2} suggests that the best-performing acquisition functions are LCB-LW and IVR-LWBO, solidifying the utility of the likelihood ratio in Bayesian optimization.  Figure \ref{fig:2} also makes clear that there is no ``warm-up'' period for the LW acquisition functions.  That the latter ``zero in'' much faster than the competition is invaluable since the power of Bayesian optimization lies in keeping the number of black-box evaluations at a minimum.

We have investigated how computation of the likelihood ratio affects algorithm efficiency (\ref{app:5}).  We have found that the benign increase in runtime associated with Algorithm \ref{alg:2}  a) can be largely mitigated if sampling of the posterior mean $\mu(\mathbf{x})$ is done frugally, and b) is inconsequential when each black-box query takes hours or days, as in most practical applications.  

\begin{figure}[!htb]
\centering
\subfigure[\label{fig:2a} 2-D Ackley function]{\includegraphics[width=3.2in]{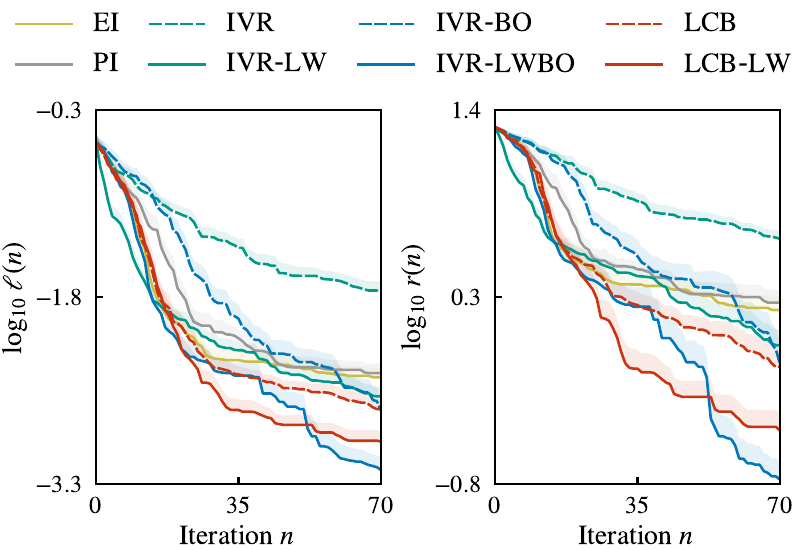}} 

\subfigure[\label{fig:2b} Branin function]{\includegraphics[width=3.2in, clip=true, trim=0 0 0 25]{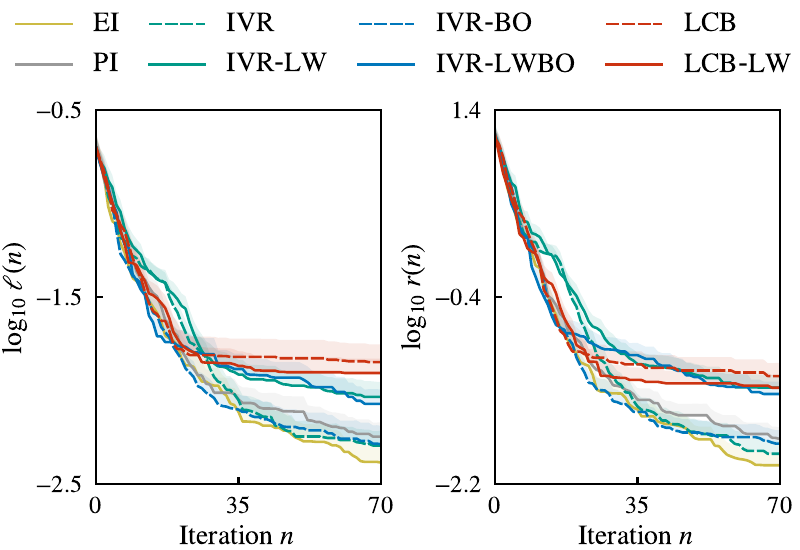}}

\subfigure[\label{fig:2c} Bukin function]{\includegraphics[width=3.2in, clip=true, trim=0 0 0 25]{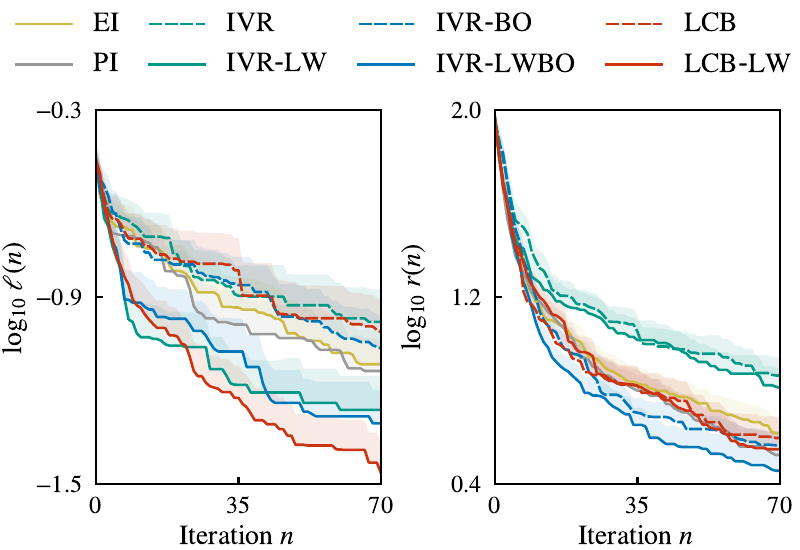}} 
\caption{For caption, see next page.}
\end{figure}

\begin{figure}[!htb]
\ContinuedFloat
\centering
\addtocounter{subfigure}{+3}
\subfigure[\label{fig:2d} 2-D Michalewicz function]{\includegraphics[width=3.2in]{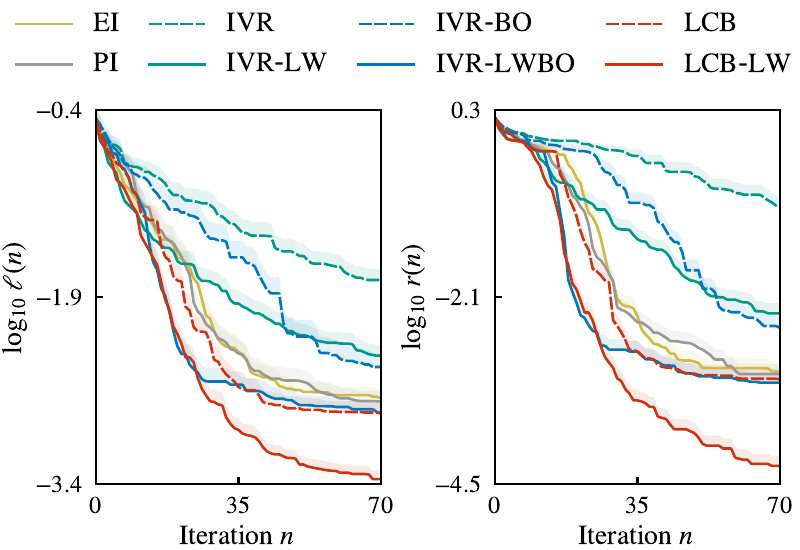}}

\subfigure[\label{fig:2e} 6-D Hartmann function]{\includegraphics[width=3.2in, clip=true, trim=0 0 0 25]{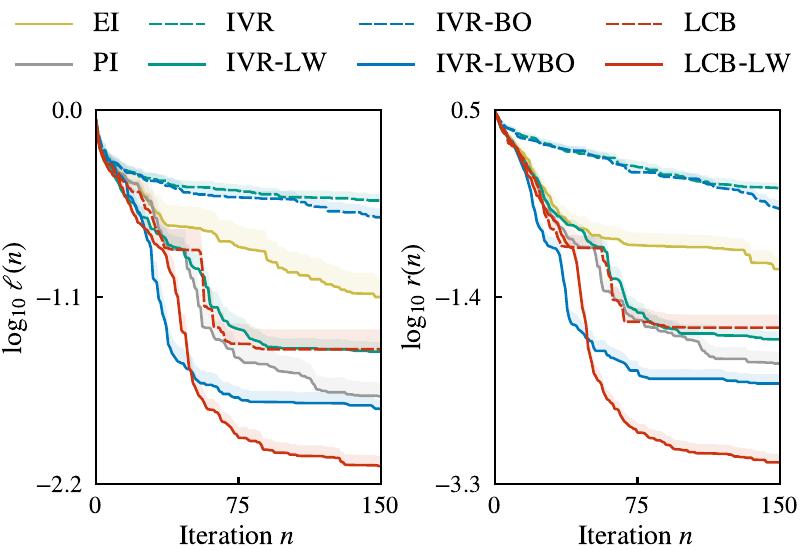}}

\subfigure[\label{fig:2f} 10-D Michalewicz function]{\includegraphics[width=3.2in, clip=true, trim=0 0 0 25]{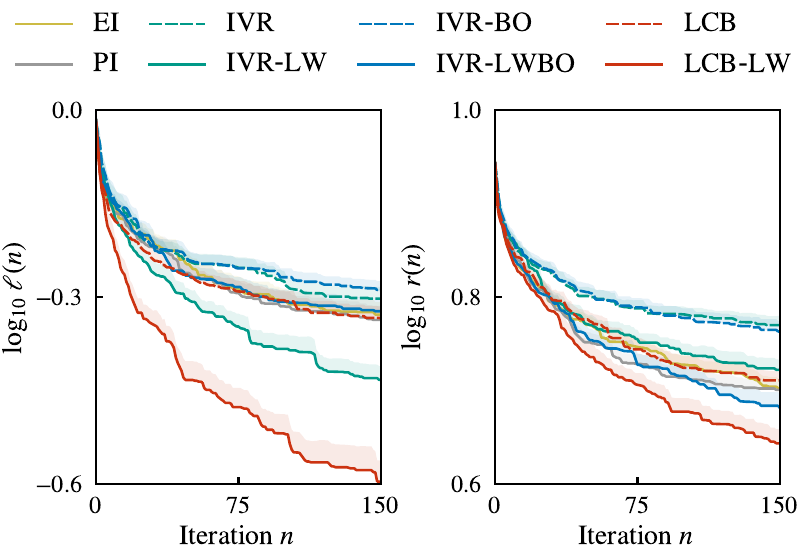}}
\caption{Performance of EI, PI, IVR(-BO), IVR-LW(BO) and LCB(-LW) for six benchmark test functions.}
\label{fig:2}
\end{figure}

\subsection{Computation of Precursors for Extreme Events in Dynamical Systems}
\label{sec:43}

For a real-world application, we consider the problem of predicting the occurrence of extreme events in dynamical systems.  This is a topic worthy of investigation because extreme events (e.g., earthquakes, road accidents, wildfires) have the potential to cause significant damage to people, infrastructure and nature \cite{albeverio2006extreme,li2011efficient,farazmand2017reduced}.  From a prediction standpoint, the central issue is to identify \textit{precursors}, i.e., those states of the system which are most likely to lead to an extreme event in the near future.  But searching for precursors is no easy task because extreme events often arise in highly complex dynamical systems, which adds to the issue of low frequency of occurrence.  To combat this, one can use Bayesian optimization to parsimoniously probe the state space of the system and thus identify ``dangerous'' regions with as little data as possible.  

Formally, the dynamical system is treated as a black box, which assigns to an initial condition $\mathbf{x}_0$ a measure of dangerousness, e.g.,
\begin{align}
F \colon  \mathbb{R}^d 	&\longrightarrow \mathbb{R} \nonumber \\
			  	\mathbf{x}_0  &\longmapsto    \max_{t\in[0, \tau]} G(S_t(\mathbf{x}_0)).
\label{eq:20}
\end{align}
Here, $t$ denotes the time variable, $S_t$ the flow map of the system (i.e., the dynamics of the black box), $G : \mathbb{R}^d \longrightarrow \mathbb{R} $ the observable of interest, and $\tau$ the time horizon over which prediction is to be performed.  In words, $F$ records the maximum value attained by the observable $G$ during the time interval $[0, \tau]$ given initial condition $\mathbf{x}_0$.  The role of Bayesian optimization is to search for those $\mathbf{x}_0$ that give rise to large values of $F$ indicative of an extreme event occurring within the next $\tau$ time units.

In practice, not the whole space of initial conditions is explored by the algorithm, as this would allow sampling of unrealistic initial conditions.  Instead, one may view extreme events as excursions from a ``background'' attractor \cite{farazmand2017variational}, for which a low-dimensional representation can be constructed by principal component analysis (PCA).  Searching for precursors in the PCA subspace is recommended when $d$ is unfathomably large (e.g., when $\mathbf{x}$ arises from discretizing a partial differential equation).  Another advantage is that the PCA subspace comes equipped with the Gaussian prior $p_\mathbf{x}(\mathbf{x}) = \mathcal{N}(\mathbf{x};\mathbf{0}, \mathbf{\Lambda})$, where the diagonal matrix $\mathbf{\Lambda}$ contains the PCA eigenvalues.  

\begin{figure}[!htb]
\addtocounter{subfigure}{-6}
\centering
\subfigure[\label{fig:3a} Trajectory in phase space]{\includegraphics[width=1.5in]{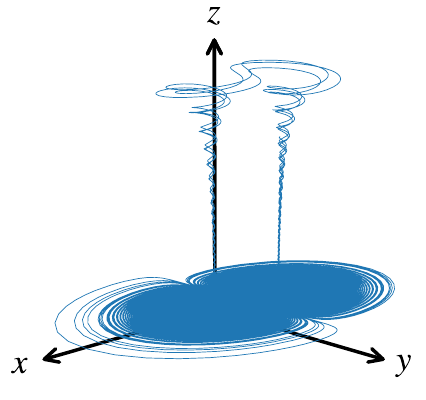}} ~~
\subfigure[\label{fig:3b} Time series of $z$ coordinate]{\includegraphics[width=2.6in]{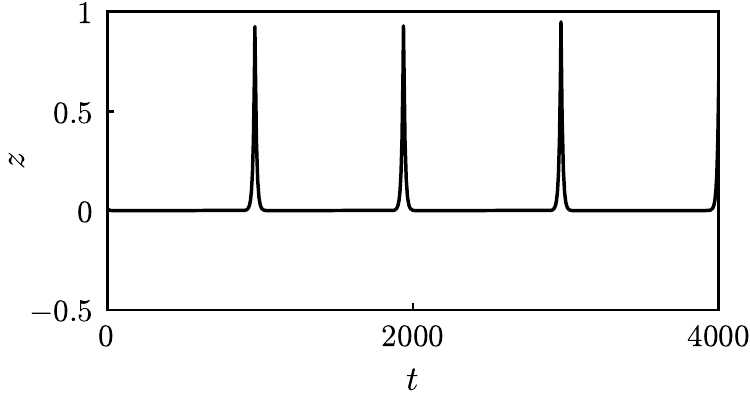}}
\caption{Dynamical system with extreme events.}
\label{fig:3}
\end{figure}

We consider the dynamical system introduced by \citet{farazmand2016dynamical} in the context of extreme-event prediction in turbulent flow.  The governing equations are given by 
\begin{subequations}
\begin{gather}
\dot{x} = \alpha x + \omega y + \alpha x^2 + 2 \omega x y +z^2, \label{eq:18a}\\
\dot{y} = -\omega x + \alpha y - \omega x^2 + 2 \alpha x y, \label{eq:18b}\\
\dot{z} = -\lambda z - (\lambda + \beta)xz, \label{eq:18c}
\end{gather}%
\end{subequations}%
with parameters $\alpha=0.01$, $\omega = 2\pi$, and $\lambda=\beta=0.1$.  (This is an example of a Shilnikov system operated backward in time \cite{wiggins1988global}.)  Figure \ref{fig:3a} shows that the system features successive ``cycles'' during which a trajectory initiated close to the origin spirals away towards the point $(-1,0,0)$, only to find itself swiftly repelled from the $z = 0$ plane.  After some ``hovering about'', the trajectory ultimately heads back to the origin, and the cycle repeats itself.  Here, extreme events correspond to ``bursts'' in the $z$ coordinate, as shown in figure \ref{fig:3b}.  

To identify precursors for these bursts, we apply Bayesian optimization to the function $-F$ with observable $G=\mathbf{e}_z^\mathsf{T}$.  The background attractor is approximated with the leading two principal components, which roughly span the $(x,y)$ plane.  We use $\tau=50$ for the prediction horizon, $\sigma_\varepsilon^2=10^{-3}$ for the noise variance, and two Gaussian mixtures for the GMM fit.  The search algorithm is initialized with three points sampled from an LHS design.  To avoid the sampling of unrealistic initial conditions, we require that $\mathbf{x}$ lie no further than four PCA standard deviations in any direction.

\begin{figure}[!htb]
\centering
\includegraphics[width=3.2in]{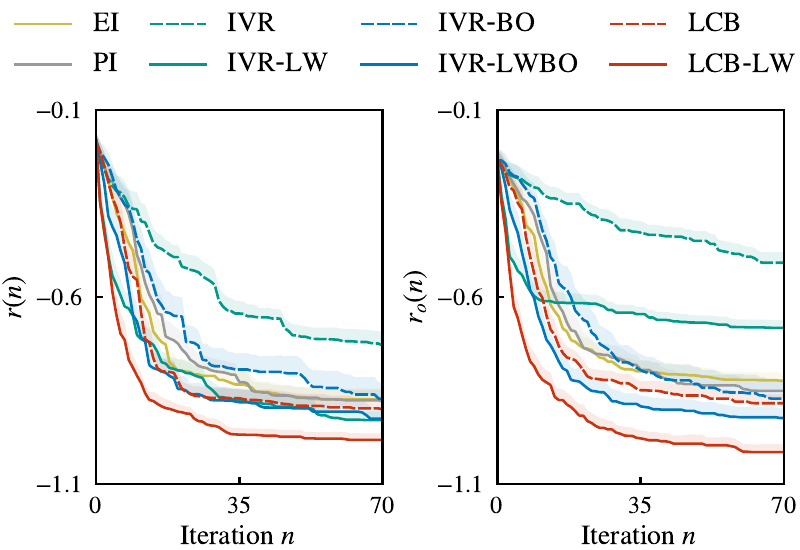}
\caption{Performance of EI, PI, IVR(-BO), IVR-LW(BO) and LCB(-LW) for computation of extreme-event precursor.}
\label{fig:4}
\end{figure}

Figure \ref{fig:4} shows that here again the LW acquisition functions trumps their unweighted cousins by a significant margin.  The two best-performing acquisition functions are again LCB-LW and IVR-LWBO, with the former predicting an objective value (about $0.98$ in figure \ref{fig:4}) within a few percent of the observation (about $0.95$ in figure \ref{fig:3b}).  We note that use of the Gaussian prior for $p_\mathbf{x}$ may lead to even greater gains when the dynamical system features multiple dangerous regions, with some more ``exotic'' than others.  We also note that uncertainty in observations (encapsulated in $\sigma_\varepsilon^2$) may be interpreted as feedback from unresolved scales in a turbulent flow, or imperfections in the experimental apparatus.  In an experiment, the question of localization noise (i.e., the error in estimating or realizing a particular state $\mathbf{x}$) is also highly relevant (see \S\ref{sec:33}).

\section{Conclusions}
\label{sec:5}

We have investigated the effect of the likelihood ratio, a quantity that accounts for the importance of the output relative to the input, on the performance of Bayesian optimization algorithms.  We have shown that use of the likelihood ratio in an acquisition function can dramatically improve algorithm efficiency, with faster convergence seen in a number of synthetic test functions.  We have proposed an approximation of the likelihood ratio as a superposition of Gaussian mixtures to make the approach tractable in high dimensions.  We have successfully applied the proposed method to the problem of extreme-event prediction in complex dynamical systems.  

While in principle the proposed approach can be applied to any optimization problem, it is expected to provide the greatest gains in situations where Bayesian optimization is a reasonable candidate and the global minimum of the objective function is several standard deviations away from the remainder of the optimization landscape.  Applications of potential interest to the practitioner include prediction of extreme events and identification of associated precursors in turbulent flow \cite{lestang2020numerical}, active control of a turbulent jet for enhanced mixing \cite{zhou2020artifical}, optimal path planning of autonomous vehicles for anomaly detection in environment exploration \cite{blanchard2020informative}, and hyper-parameter training of deep-learning algorithms \cite{snoek2012practical}.

\section*{Acknowledgments}
The authors acknowledge support from the Army Research Office (Grant No. W911NF-17-1-0306) and the 2020 MathWorks Faculty Research Innovation Fellowship.

\section*{References}
\bibliography{mybibfile}

\appendix 

\section{Analytical Expressions for the IVR Acquisition Function with RBF Kernel}
\label{app:1}

We first rewrite the formula for IVR using the GP expression for the posterior covariance:
\begin{subequations}
\begin{align}
\sigma^2(\mathbf{x})\, a_\textit{IVR}(\mathbf{x}) &= \int \left[k(\mathbf{x},\mathbf{x}') -k(\mathbf{x},\mathbf{X}) \mathbf{K}^{-1} k(\mathbf{X},\mathbf{x}')\right]^2  \, \mathrm{d}\mathbf{x}' \label{eq:a1a} \\
&= \int k(\mathbf{x}, \mathbf{x}') k(\mathbf{x}', \mathbf{x}) \, \mathrm{d}\mathbf{x}' \nonumber \\
&\qquad + k(\mathbf{x}, \mathbf{X}) \mathbf{K}^{-1} \left[ \int k(\mathbf{X}, \mathbf{x}') k(\mathbf{x}', \mathbf{X}) \, \mathrm{d}\mathbf{x}' \right] \mathbf{K}^{-1}k( \mathbf{X}, \mathbf{x})   \nonumber \\
&\qquad - 2 k(\mathbf{x}, \mathbf{X}) \mathbf{K}^{-1} \int k(\mathbf{X}, \mathbf{x}') k(\mathbf{x}', \mathbf{x})  \, \mathrm{d}\mathbf{x}'.
\label{eq:a1b}%
\end{align}%
\end{subequations}
If we introduce
\begin{equation}
\hat{k}(\mathbf{x}_1, \mathbf{x}_2) = \int k(\mathbf{x}_1, \mathbf{x}') k( \mathbf{x}', \mathbf{x}_2) \mathrm{d}\mathbf{x},
\label{eq:a2}
\end{equation}
then (\ref{eq:a1b}) can be rewritten as
\begin{align}
\sigma^2(\mathbf{x})\, a_\textit{IVR}(\mathbf{x}) &= \hat{k}(\mathbf{x}, \mathbf{x}) +  k(\mathbf{x}, \mathbf{X}) \mathbf{K}^{-1} \left[ \hat{k}(\mathbf{X}, \mathbf{X}) \mathbf{K}^{-1}k( \mathbf{X}, \mathbf{x}) - 2 \hat{k}(\mathbf{X},  \mathbf{x}) \right]\!.
\label{eq:a3}%
\end{align}%
This helps us realize that to compute IVR and its gradients, we only need a mechanism to compute (\ref{eq:a2}) and its gradients, regardless of the choice of GP kernel.  

For the RBF kernel
\begin{equation}
k(\mathbf{x},\mathbf{x}'; \mathbf{\Theta}) = \sigma_f^2 \exp \!\left[ -(\mathbf{x} - \mathbf{x}')^\mathsf{T} \mathbf{\Theta}^{-1}(\mathbf{x} - \mathbf{x}') /2\right],
\label{eq:s14}
\end{equation}
we have
\begin{subequations}
\begin{equation}
\hat{k}(\mathbf{x}_1, \mathbf{x}_2) = \sigma_f^2 \pi^{d/2} |\mathbf{\Theta}|^{1/2} k(\mathbf{x}_1, \mathbf{x}_2; 2 \mathbf{\Theta})
\end{equation}
and
\begin{equation}
\frac{\mathrm{d}}{\mathrm{d}\mathbf{x}_1}\hat{k}(\mathbf{x}_1, \mathbf{x}_2) = -\hat{k}(\mathbf{x}_1, \mathbf{x}_2) (\mathbf{x}_1 - \mathbf{x}_2)^\mathsf{T} (2\mathbf{\Theta})^{-1}.
\end{equation}
\end{subequations}
For further details, we refer the reader to \citet{mchutchon2013differentiating}.

\section{Mathematical Derivation of IVR-LW}
\label{app:2}

We begin with Theorem 1 of \citet{mohamad2018sequential} which states that for small enough $\sigma(\mathbf{x})$,
\begin{equation}
a_L(\mathbf{x}) \approx  \int_\mathcal{Y} \frac{1}{p_\mu(y)}  \left| \frac{\mathrm{d}}{\mathrm{d}y}\mathbb{E}[\sigma^2(\mathbf{x}'; \mathbf{x}) \cdot  \mathbf{1}_{\mu(\mathbf{x}')=y}] \right|  \mathrm{d}y,
\end{equation}
where $\mathbb{E}$ is the expectation with respect to $p_\mathbf{x}$, and $\mathcal{Y}$ is the domain over which the pdf $p_\mu$ is defined.  Standard inequalities \cite{kwong2006norm} allow us to bound the above as follows:
\begin{equation}
\int_\mathcal{Y} \left| \frac{1}{p_\mu(\mu)} \frac{\mathrm{d}}{\mathrm{d}y}\mathbb{E}[\sigma^2(\mathbf{x}'; \mathbf{x}) \cdot  \mathbf{1}_{\mu(\mathbf{x}')=y} ]\right|  \mathrm{d}y \leq K \int_\mathcal{Y} \frac{1}{p_\mu(\mu)}   \mathbb{E}[\sigma^2(\mathbf{x}'; \mathbf{x}) \cdot  \mathbf{1}_{\mu(\mathbf{x}')=y}]  \,\mathrm{d}y,
\end{equation}
where $K$ is a positive constant.  But we note that 
\begin{subequations}
\begin{align}
\mathbb{E}[\sigma^2(\mathbf{x}'; \mathbf{x}) \cdot  \mathbf{1}_{\mu(\mathbf{x}')=y}] &= \int \sigma^2(\mathbf{x}'; \mathbf{x}) \cdot  \mathbf{1}_{\mu(\mathbf{x}')=y} \,p_\mathbf{x}(\mathbf{x}') \, \mathrm{d} \mathbf{x}'\\
& = \int_{\mu(\mathbf{x}')=y} \sigma^2(\mathbf{x}'; \mathbf{x}) p_\mathbf{x}(\mathbf{x}') \, \mathrm{d} \mathbf{x}'.%
\end{align}%
\end{subequations}%
Therefore,
\begin{equation}
\int_\mathcal{Y} \frac{1}{p_\mu(\mu)}   \mathbb{E}[\sigma^2(\mathbf{x}'; \mathbf{x}) \cdot  \mathbf{1}_{\mu(\mathbf{x}')=y}]  \,\mathrm{d}y = \int_{\mu(\mathbf{x}')\in \mathcal{Y}} \frac{\sigma^2(\mathbf{x}'; \mathbf{x}) p_\mathbf{x}(\mathbf{x}')}{p_\mu(\mu(\mathbf{x}'))}    \, \mathrm{d} \mathbf{x}'.%
\label{eq:17b}%
\end{equation}%
In practice, the domain of integration in \eqref{eq:17b} is replaced with the support of the input pdf $p_\mathbf{x}$:
\begin{equation}
a_B(\mathbf{x}) = \int \sigma^2(\mathbf{x}'; \mathbf{x})  \frac{p_\mathbf{x}(\mathbf{x}')}{p_\mu(\mu(\mathbf{x}'))} \, \mathrm{d}\mathbf{x}'.
\end{equation}
Finally, it should be clear that the optimization problem
\begin{equation}
\min_{\mathbf{x} \in \mathcal{X}} \int \sigma^2(\mathbf{x}'; \mathbf{x})  \frac{p_\mathbf{x}(\mathbf{x}')}{p_\mu(\mu(\mathbf{x}'))} \, \mathrm{d}\mathbf{x}'
\end{equation}
is strictly equivalent to
\begin{equation}
\max_{\mathbf{x} \in \mathcal{X}} \int [\sigma^2(\mathbf{x}') -\sigma^2(\mathbf{x}'; \mathbf{x})] \frac{p_\mathbf{x}(\mathbf{x}')}{p_\mu(\mu(\mathbf{x}'))}  \, \mathrm{d}\mathbf{x}',
\label{eq:S222}
\end{equation}
since the term involving $\sigma^2(\mathbf{x}')$ in \eqref{eq:S222} does not depend on the optimization variable $\mathbf{x}$.  We can then rewrite the difference of variances using the trick of \citet{gramacy2009adaptive}. We thus obtain
\begin{equation}
\max_{\mathbf{x} \in \mathcal{X}} \frac{1}{\sigma^2(\mathbf{x})} \int \mathrm{cov}^2(\mathbf{x}, \mathbf{x}') \frac{p_\mathbf{x}(\mathbf{x}')}{p_\mu(\mu(\mathbf{x}'))} \, \mathrm{d}\mathbf{x}',\label{eq:S223}
\end{equation}
which concludes the derivation of IVR-LW.

\section{Analytical Expressions for the IVR-LW Acquisition Function with RBF Kernel}
\label{app:3}

With the likelihood ratio being approximated with a GMM, the IVR-LW acquisition function becomes
\begin{equation}
a_\textit{IVR-LW}(\mathbf{x}) \approx \frac{1}{\sigma^2(\mathbf{x})} \sum_{i=1}^{n_\mathit{GMM}}\beta_i \, a_i(\mathbf{x}),
\end{equation}
where each $a_i$ is given by
\begin{equation}
a_i(\mathbf{x}) = \int \mathrm{cov}^2(\mathbf{x}, \mathbf{x}') \,\mathcal{N}(\mathbf{x}';\boldsymbol{\omega}_i, \mathbf{\Sigma}_i)  \, \mathrm{d}\mathbf{x}'.
\end{equation}
Using the formula for the posterior covariance, we get
\begin{subequations}
\begin{align}
a_i(\mathbf{x}) &= \int \left[k(\mathbf{x},\mathbf{x}') -k(\mathbf{x},\mathbf{X}) \mathbf{K}^{-1} k(\mathbf{X},\mathbf{x}')\right]^2 \mathcal{N}(\mathbf{x}';\boldsymbol{\omega}_i, \mathbf{\Sigma}_i)  \, \mathrm{d}\mathbf{x}' \label{eq:s214a} \\
&= \int k(\mathbf{x}, \mathbf{x}') k(\mathbf{x}', \mathbf{x})\, \mathcal{N}(\mathbf{x}';\boldsymbol{\omega}_i, \mathbf{\Sigma}_i)  \, \mathrm{d}\mathbf{x}' \nonumber \\
&\qquad + k(\mathbf{x}, \mathbf{X}) \mathbf{K}^{-1} \left[ \int k(\mathbf{X}, \mathbf{x}') k(\mathbf{x}', \mathbf{X})\, \mathcal{N}(\mathbf{x}';\boldsymbol{\omega}_i, \mathbf{\Sigma}_i)  \, \mathrm{d}\mathbf{x}' \right] \mathbf{K}^{-1}k( \mathbf{X}, \mathbf{x})   \nonumber \\
&\qquad - 2 k(\mathbf{x}, \mathbf{X}) \mathbf{K}^{-1} \int k(\mathbf{X}, \mathbf{x}') k(\mathbf{x}', \mathbf{x}) \, \mathcal{N}(\mathbf{x}';\boldsymbol{\omega}_i, \mathbf{\Sigma}_i)  \, \mathrm{d}\mathbf{x}', \\
&= \hat{k}_i(\mathbf{x}, \mathbf{x})  +  k(\mathbf{x}, \mathbf{X}) \mathbf{K}^{-1} \left[  \hat{k}_i(\mathbf{X}, \mathbf{X}) \mathbf{K}^{-1}k( \mathbf{X}, \mathbf{x}) - 2  \hat{k}_i(\mathbf{X}, \mathbf{x}) \right]\!,
\label{eq:s214b}%
\end{align}%
\end{subequations}
where we have defined 
\begin{equation}
\hat{k}_i(\mathbf{x}_1, \mathbf{x}_2) = \int k(\mathbf{x}_1, \mathbf{x}') k( \mathbf{x}', \mathbf{x}_2) \, \mathcal{N}(\mathbf{x}';\boldsymbol{\omega}_i, \mathbf{\Sigma}_i)  \, \mathrm{d}\mathbf{x}'.
\label{eq:s213}
\end{equation}
Therefore, to evaluate $a_i$ and its gradients, we only need a mechanism to compute $\hat{k}_i$ and its gradients.  For the RBF kernel \eqref{eq:s14}, it is straightforward to show that 
\begin{subequations}
\begin{equation}
\hat{k}_i(\mathbf{x}_1, \mathbf{x}_2) = |2 \mathbf{\Sigma}_i \mathbf{\Theta}^{-1} + \mathbf{I}|^{-1/2} k(\mathbf{x}_1, \mathbf{x}_2; 2 \mathbf{\Theta}) k(\mathbf{x}_1 + \mathbf{x}_2, \boldsymbol{\omega}_i; \mathbf{\Theta} + 2 \mathbf{\Sigma}_i)
\label{eq:s215}
\end{equation}
and
\begin{equation}
\frac{\mathrm{d}\hat{k}_i}{\mathrm{d}\mathbf{x}_1}  = \hat{k}_i(\mathbf{x}_1, \mathbf{x}_2) \left\{ -\mathbf{x}_1^\mathsf{T} \mathbf{\Theta}^{-1} + \frac{1}{2}\left[\boldsymbol{\omega}_i^\mathsf{T} + (\mathbf{x}_1+\mathbf{x}_2)^\mathsf{T}\mathbf{\Theta}^{-1} \mathbf{\Sigma}_i \right] (  \mathbf{\Sigma}_i + \mathbf{\Theta} / 2 )^{-1}\right\}\!.
\label{eq:s216}
\end{equation}
\end{subequations}
For further details, we refer the reader to \citet{mchutchon2013differentiating}.

\section{Analytical Expressions for Benchmark Test Functions}
\label{app:4}

\begin{figure}[!htb]
\centering
\subfigure[2-D Ackley function]{\includegraphics[width=2in]{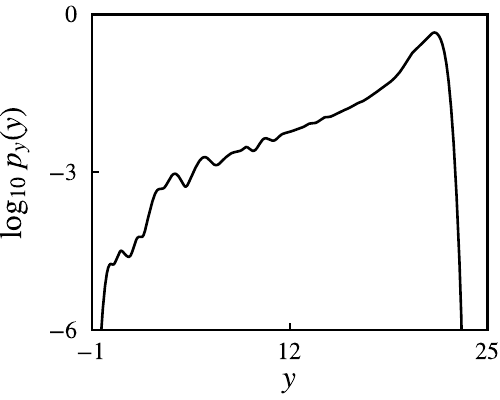}} \quad
\subfigure[Branin function]{\includegraphics[width=2in]{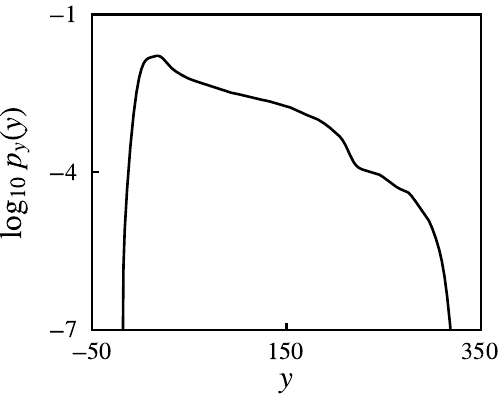}} 

\subfigure[Bukin function]{\includegraphics[width=2in]{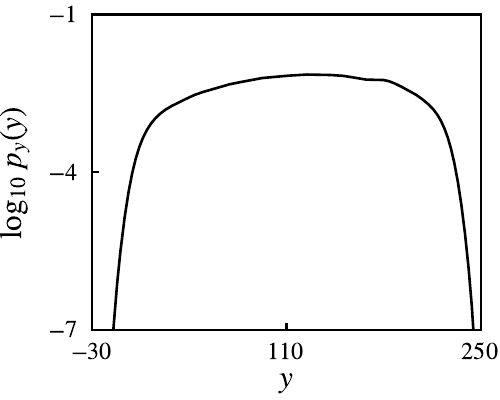}} \quad
\subfigure[2-D Michalewicz function]{\includegraphics[width=2in]{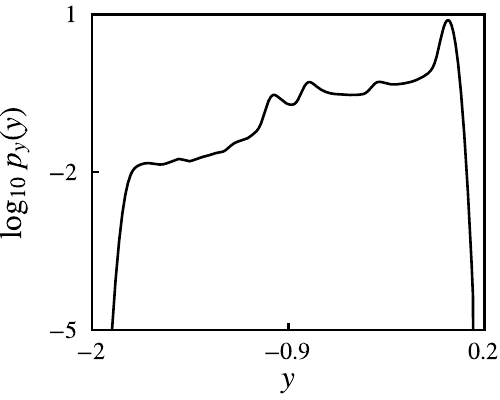}} 

\subfigure[6-D Hartmann function]{\includegraphics[width=2in]{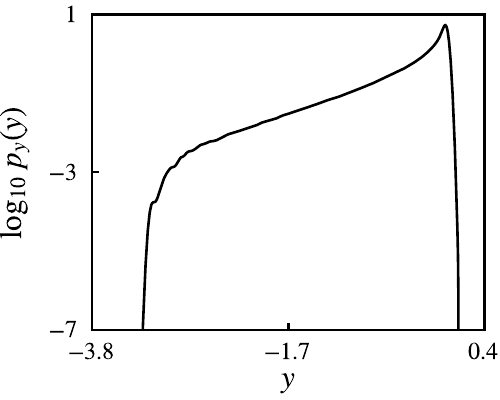}} \quad
\subfigure[10-D Michalewicz function]{\includegraphics[width=2in]{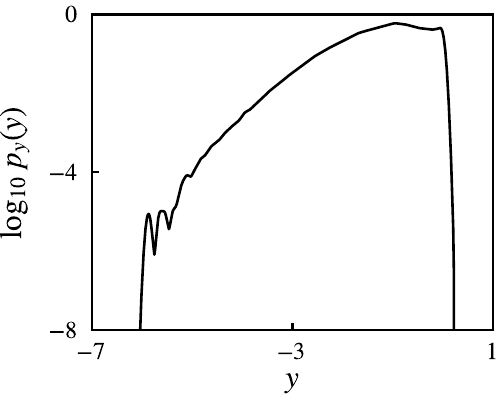}}
\caption{Conditional pdf with uniform input for the six benchmark test functions considered in this work.}
\label{fig:S1}
\end{figure}

The analytical expressions for the synthetic test functions considered in Section 4.2 are given below.  Further details may be found in \citet{simulationlib} and \citet{jamil2013literature}.  For these test functions, figure \ref{fig:S1} shows the conditional pdf of the output for uniformly distributed input.  We used $10^{5}$ samples for the 2-D functions, and $10^{6}$ samples for the 6-D Hartmann and 10-D Michalewicz functions.

\textbf{Ackley function:}
\begin{equation}
f(\mathbf{x}) = -a \exp \! \left[ -b \sqrt{\frac{1}{d}\sum_{i=1}^d x_i^2} \right] - \exp\! \left[\frac{1}{d} \sum_{i=1}^d \cos(c x_i) \right] +a +\exp(1),
\end{equation}
where $d$ denotes the dimensionality of the function, and $a=20$, $b=0.2$, and $c=2\pi$.  

\textbf{Branin function:}
\begin{equation}
f(\mathbf{x}) = a(x_2 - b x_1^2 + c x_1 -r)^2 + s(1-t) \cos(x_1) + s,
\end{equation}
where $a=1$, $b=5.1/(4\pi^2)$, $c=5$, $r=6$, $s=10$, and $t=1/(8\pi)$.  

\textbf{Bukin function:}
\begin{equation}
f(\mathbf{x}) = 100 \sqrt{|x_2-0.01x_1^2|} + 0.01 |x_1+10|
\end{equation}

\textbf{Michalewicz:}
\begin{equation}
f(\mathbf{x}) = - \sum_{i=1}^d \sin(x_i) \sin^{2m}(ix_i^2/\pi),
\end{equation}
where $d$ denotes the dimensionality of the function, and $m$ controls the steepness of the valleys and ridges.  In this work we use $m=10$, making optimization extremely challenging.  

\textbf{6-D Hartmann function:}
\begin{equation}
f(\mathbf{x}) = - \sum_{i=1}^4 a_i \exp\!\left[-\sum_{j=1}^6 A_{ij}(x_j-P_{ij})^2 \right]\!,
\end{equation}
where
\begin{subequations}
\begin{gather}
\mathbf{a} = \begin{bmatrix} 1 & 1.2 & 3 & 3.2 \end{bmatrix}^\mathsf{T}, \\
\mathbf{A} =  \begin{bmatrix} 	10 	& 3 	& 17 	& 3.5	  & 1.7 & 8 \\
          					0.05 	& 10	& 17	& 0.1	  & 8	   & 14 \\
						3 	& 3.5	& 1.7	& 10	  & 17  & 8 \\ 
						17 	& 8	& 0.05& 10 & 0.1 & 14 
						\end{bmatrix},\\
\mathbf{P} =  \begin{bmatrix} 	0.1312 & 0.1696 & 0.5569 & 0.0124 & 0.8283 & 0.5886 \\
                             0.2329 & 0.4135 & 0.8307 & 0.3736 &0.1004 & 0.9991 \\
                             0.2348 & 0.1451 & 0.3522 & 0.2883 & 0.3047 & 0.6650 \\
                             0.4047 & 0.8828 & 0.8732 & 0.5743 & 0.1091 & 0.0381
						\end{bmatrix}.
\end{gather}
\end{subequations}

\section{Comparison of Runtime for Likelihood-Weighted and Unweighted Acquisition Functions}
\label{app:5}

To investigate how computation of the likelihood ratio affects the overall efficiency of the algorithm, we proceed as follows.  For the Ackley function, we consider an initial dataset composed of ten LHS input--output pairs and, for the eight acquisition functions considered in Section 4.2, record the time required to perform a single iteration of the Bayesian algorithm.  During that time interval, the following operations are performed: computation of the likelihood ratio (for LW acquisition functions only), minimization of the acquisition function over the input space, query of the objective function at the best next point, and update of the surrogate model.

Since the focus is on the likelihood ratio, we investigate the effect of the following parameters on runtime: number of samples $n_\textit{samples}$ drawn from the posterior mean used in KDE, number of Gaussian mixtures $n_\textit{GMM}$ used in the GMM fit, and dimensionality of the objective function $d$.  The baseline case uses $n_\textit{samples}=10^{6}$, $n_\textit{GMM}=2$, and $d=2$.  (These are the parameters used to generate the results in figure 2(a).)  For each parameter, we perform 100 experiments (each with a different LHS initialization, followed by a single Bayesian iteration) and report the median runtime (in seconds) for EI, PI, IVR(-BO), IVR-LW(BO), and LCB(-LW).

\begin{figure}[!htb]
\centering
\includegraphics[width=4.8in]{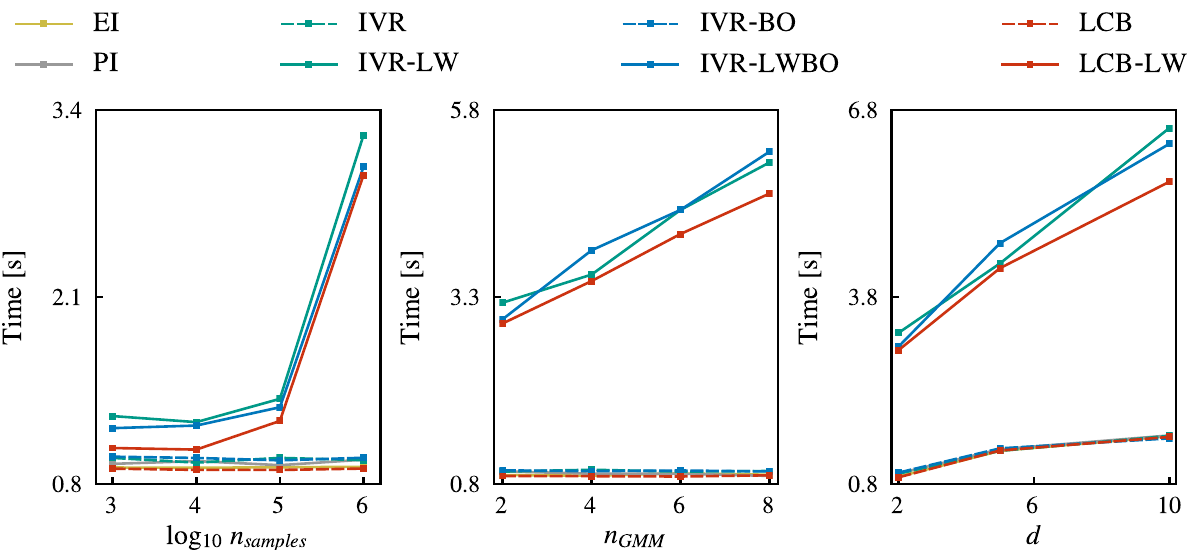}
\caption{Effect of $n_\textit{samples}$, $n_\textit{GMM}$, and $d$, on single-iteration runtime for the Ackley function.}
\label{fig:S2}
\end{figure}

Figure \ref{fig:S2} shows that computation of the likelihood ratio has a mildly adverse effect on algorithm efficiency, with the main culprit being the sampling of the posterior mean.  The left panel in figure \ref{fig:S2} shows a relatively strong dependence of the runtime on $n_\textit{samples}$.  We have verified that for the LW acquisition functions, very little time is spent in the KDE part of the algorithm---as discussed in Section 3.3, one-dimensional FFT-based KDE scales linearly with the number of samples.  Instead, most of the iteration time is spent in generating the samples of the posterior mean $\mu(\mathbf{x})$.  Little can be done to avoid this issue, except for using a reasonably small number of samples.  For $n_\textit{samples} < 10^{5}$, the left panel in figure \ref{fig:S2} shows that runtime for LW acquisition functions is not dramatically larger than that for unweighted acquisition functions.  We note that for relatively low dimensions, reducing the number of samples has virtually no effect on the accuracy of the KDE.  For $n_\textit{samples}=10^{6}$, the center and right panels in figure \ref{fig:S2} suggest that runtime scales linearly with the number of Gaussian mixtures and dimensionality, which is unproblematic from the standpoint of efficiency.  

As a final word, we note that the runtimes for LW acquisition functions remain on the same order of magnitude as that for unweighted acquisition functions.  In practical applications, evaluation of the black-box objective function may take days or even weeks, so whether computation and optimization of the acquisition function takes one or ten seconds is inconsequential.

\end{document}